# A Differentiable Framework for General Circulation Model Precipitation Bias Correction

Kamlesh Sawadekar[1], Seth McGinnis[2], Peijun Li[1], Kathryn Lawson[1], Chaopeng Shen[1]

[1]Department of Civil and Environmental Engineering, The Pennsylvania State University, University Park, Pennsylvania 16801 US
[2]Computational & Information Systems Laboratory, National Center for Atmospheric Research, Boulder, Colorado 80307 US

*Correspondence to*: Chaopeng Shen (cshen@engr.psu.edu)

**Abstract.** Systematic biases in General Circulation Model (GCM) outputs limit their direct applicability in regional planning, making bias correction a technically demanding but necessary step for both short-term and long-term impact assessment. Correcting precipitation is particularly challenging due to its non-Gaussian distribution, intermittent nature, and heavy-tailed extremes. However, traditional statistical bias-correction methods have limited ability to learn systematic patterns from large datasets or generalize to new locations. While machine learning (ML) provides greater flexibility, it can produce unpredictable and difficult-to-interpret results, limiting generalization across GCMs and locations. In this study, we propose a differentiable bias-adjustment framework called δCLIMBA, or dCLIMBA, that learns a spatiotemporally adaptive parametric bias-adjustment procedure, rather than corrected precipitation directly, between historical CMIP6 model outputs and a gridded observation-based dataset, Livneh. Results demonstrate that the proposed method corrects the magnitude and distribution of extreme precipitation with particularly strong performance in the upper tail. The quantile distribution of precipitation was well reproduced across diverse U.S. cities, and spatial patterns were comparable to those from the widely used LOCA2 statistical downscaling product. In addition, the framework showed partial future trend preservation and promising attenuation of marginal biases in unseen regions. This work presents a modular and efficient bias-correction approach. The differentiable approach provides an easy-to-use option for connecting atmospheric-model outputs to on-the-ground impacts.

## 1. Introduction

General Circulation Models (GCMs) are essential tools for projecting future atmospheric regime change, but their outputs are subject to systematic biases that limit their direct use in regional impact assessments. These models simulate interactions between the atmosphere, hydrosphere, cryosphere, and biosphere, and are used by policymakers and scientists to develop mitigation and adaptation strategies (Intergovernmental Panel On Climate Change, 2014). Our research focuses on simulations from the Coupled Model Intercomparison Project (CMIP), which produces centennial climate projections from multiple GCMs (Eyring et al., 2016a). Fundamental insights from these simulations guide policymakers in the fields of water, energy, and biodiversity, but their applicability for local-scale impact studies is hindered by their coarse spatial resolution and intrinsic biases (Palmer & Stevens, 2019). These systematic biases, which critically limit the direct use of GCMs in regional impact assessments, stem from multiple factors, including structural errors and simplifications, inaccurate parameter values, and uncertainties in forcings and boundary conditions. Approximations of sub-grid-scale processes such as cloud formation, convection, and land-atmosphere interactions create errors that may be negligible at the GCM scale (Hourdin et al., 2017), but result in systematic biases when disaggregated or interpolated to finer scales; these biases can be particularly troublesome in regions with complex topography or heterogeneous land surface properties where coarse-scale GCMs struggle to capture local variability (Giorgi & Mearns, 2002).

To enhance the applicability of GCMs on the finer scales, laborious and technically demanding bias correction techniques (also referred to as bias adjustment) have often been used to align GCM outputs with chosen observational datasets. Bias adjustment is traditionally carried out by methods like quantile mapping, linear scaling, and regression-based transfer functions (Piani et al., 2010). Though these methods are very popular, their utility is impaired by their dependence on linear assumptions and inability to capture complex, non-linear dependencies between atmospheric variables (Ehret et al., 2012). A major limitation of these methods is that they usually depend on the assumption of a stationary relationship between model outputs and observations. As atmospheric dynamics change over time, the calibrations learned on historical data become inaccurate for future conditions, leading to errors in extreme event projections (Deser et al., 2012). Traditional methods are also applied on a point-by-point basis with no regard for spatial coherence, but maintaining realistic spatial patterns is also important, e.g., for understanding storm clustering (Hess et al., 2023; Lovejoy et al., 1987). In addition, unless special measures are taken, adjusting bias based on stationary relationships can distort the simulation generated by the model (Buser et al., 2009; Ehret et al., 2012).

Machine learning (ML)-based bias correction has recently progressed beyond traditional quantile mapping toward architectures explicitly tailored to correcting errors (Hess et al., 2023; Pan et al., 2021; F. Wang et al., 2023). For example, Hess et al. (2023) used an adversarial domain adaptation framework to learn a mapping between historical GCM precipitation and observations that corrects biases while preserving spatiotemporal coherence. Pan et al. (2021) demonstrated that generative adversarial networks can improve daily precipitation fields from CMIP6-class Earth system models, reducing systematic errors in means, extremes, and spatial patterns compared to standard statistical corrections. Wang et al. (2023) developed a customized convolutional neural network that incorporates physically relevant covariates, multitask learning, and weighted loss functions, showing enhanced skill relative to conventional approaches in downscaling and bias-correcting hourly precipitation, particularly for extreme events. Together, these ML-based frameworks illustrate the potential of flexible, data-driven bias correction of GCM outputs.

Despite these successes, purely data-driven ML approaches face challenges like generalization and interpretability. Like their statistical counterparts, ML approaches are trained on historical datasets, and hence are prone to overfitting and may fail to generalize across multiple GCMs. They can produce outputs in opaque ways — it is unclear what exact transformations they apply, when they are reliable, and whether they have failed silently. ML training can result in the disruption of overall physical consistency due to the lack of physical or (in case of precipitation) hydrological constraints (Harder et al., 2022; Maraun et al., 2010). Their "black boxes" nature makes it difficult to ensure that their outputs align with established physical principles (Harder et al., 2022). ML models also require large amounts of training data, which can be scarce or inconsistent in certain regions, reducing their robustness for global applications.

To obtain an interpretable bias corrector, we adopt differentiable modeling frameworks, which integrate prescribed equations with data-driven models. “Differentiability” refers to the computing paradigm that, with any method, rapidly and accurately produces gradients to achieve the large-scale optimization of the combined system (Shen et al., 2023). These frameworks allow for end-to-end training of neural networks that are embedded into a system that adheres to physical principles such as conservation of energy and mass (Tsai et al., 2021). Differentiable programming has already demonstrated success in hydrological modeling and ecological simulations, offering a pathway to more interpretable and generalizable models (Aboelyazeed et al., 2023; Baste et al., 2025; Bindas et al., 2024; Fang & Gentine, 2024; Feng et al., 2022; Huynh et al., 2025; P. Li et al., 2024; Z. Li et al., 2024; Rahmani et al., 2023). Such frameworks have already shown potential in understanding and mitigating some biases in reanalysis datasets through parametric fusion informed by hydrology (Sawadekar et al., 2025).

Here, we develop a Differentiable CLImate Model Bias Adjustment (δCLIMBA) model for daily precipitation simulated by GCMs. This method applies a parameterized transformation optimized over matching quantiles. This framework can be applied to any GCM and observational dataset and the trained neural network can be easily applied in space and time. There is the potential to study the model biases it corrects. The proposed framework improves bias adjustment of GCM outputs for variables including precipitation, which is one of the most difficult variables to bias correct or downscale due to its intermittent nature and highly skewed, non-Gaussian distribution (Hess & Boers, 2021). By leveraging large-scale gridded observation datasets, this framework learns spatial and temporal patterns of bias while incorporating the requisite constraints on the outputs. We will also leverage the differentiable framework to bias correct precipitation in unseen regions (those without extensive ground truth data) and learn about its capability to generalize spatially since we explicitly integrate topographical attributes into our model.

To achieve this, the study focuses on the following research questions:

- *Can the framework generalize across space and across GCMs while preserving trends?*
- *How well does the framework attenuate marginal biases in intensity, extremes, and threshold exceedances?*
- *Does the framework distort spatial patterns in precipitation?*

## 2 Methods and Data

We used two jointly connected temporal and spatial encoders to predict coefficients for a monotonic softplus transformation of precipitation from GCM outputs, with the GCM itself and a small set of attributes serving as model inputs (Fig. 1). This transformation is 'learned' against a quantile-based loss function (along with few more complimentary loss functions). We evaluate the overall utility of the bias-adjusted GCM output using a number of ETCCDI precipitation indices.

### 2.1 Differentiable Bias Adjustment Model (δCLIMBA)

The bias adjustment here uses a differentiable model framework inspired by Feng et al. (2022) and Li et al. (2025), wherein a neural network is connected to a process-based model to improve parameterization. The philosophy behind this methodology is to parameterize the transformation of GCM precipitation by optimizing the match between quantiles, with the aim of learning interpretable and generalizable mappings that can be ported to untrained locations.

The δCLIMBA model comprises two neural networks: $NN_1$ and $NN_2$. $NN_1$ is a temporal encoder which is responsible for learning biases based on temporal patterns. We evaluated several candidates for $NN_1$ including a simple Multi-Layer Perceptron (MLP) (P. Li et al., 2020; Rumelhart et al., 1986), a Long Short-Term Memory (LSTM) model (Feng et al., 2020; Hochreiter & Schmidhuber, 1997), and a one-dimensional Convolutional Neural Network (CNN-1D) (Feng et al., 2021; LeCun et al., 1989; P. Li et al., 2024) that would generate parameters to facilitate the GCM's transformation. $NN_2$ is an attention-based (Vaswani et al., 2017) spatial encoder that learns spatial relationships between neighboring inputs. The spatial self-attention is performed independently at each time step over the nodes within a patch, where queries, keys, and values are derived from node features and augmented with a learnable, head-specific offset that encodes pairwise geodesic relationships (relative displacement, distance, and bearing). This geometry-aware formulation lets the model weight neighboring nodes according to physically meaningful spherical geometry rather than fixed or isotropic spatial kernels, akin to geometry-aware and geodesic self-attention mechanisms proposed for point clouds and images (Guo et al., 2020; Miyato et al., 2023). For each grid cell, we chose the 16 nearest neighbors whose time series are positively correlated with that of the target grid cell. Our final model configuration uses a CNN-1D with 2 hidden layers of size 64 for $NN_1$, and a Transformer with a self-attention mechanism employing 2 attention heads and 64 model (hidden) dimensions for $NN_2$. LSTM can be used for $NN_1$ instead, though we went with the CNN-1D. We trained the model using the Adam optimizer (Kingma & Ba, 2017) with a learning rate at 1e-4. The batch size was chosen to be 5, and the time series length was 365 (days). This configuration was chosen based on overall performance across different precipitation indices. The full process can be mathematically described as:

$$NN_{BA}(X, X_{t-1}, X_{t-2}, X_{t-3}, A, I_t) = [\alpha, w, s, b, c] \equiv \theta \tag{1}$$

$$X_{BA} = \alpha x + \left( \sum_{z=1}^{Z} w_z softplus\big(s_z(x - b_z)\big) \right) + c \tag{2}$$

Here the bias adjustment neural network, $NN_{BA}$, is the jointly connected network of $NN_1$ and $NN_2$. X is the matrix of GCM precipitation with dimensions (batch size: 5, neighbors: 16, time: *t* (1-365)). A is the list of static attributes that could logically influence precipitation bias, such as elevation, slope, aspect, and landcover. *It* is a wet day indicator where $I_t = 1$ if daily precipitation (x) ≥ 1 and $I_t = 0$ otherwise. The output of $NN_{BA}$ is a set (θ) of temporally- and spatially varying parameters ($\alpha$, $w_z$, $s_z$, $b_z$, c) for the 'softplus monotone basis' mapping, which enforces monotonicity through softplus-based basis bumps (Zhou, 2016). We use the softplus function (Dugas et al., 2001) to ensure positivity while maintaining smooth gradients, avoiding optimization issues associated with non-differentiable or saturating alternatives. *Z* is the number of basis bumps (a hyperparameter, tuned to 8 here). $X_{BA}$ is the resulting bias-corrected GCM precipitation.

We used the following composite loss function in training:

$$L = p_1 Q + p_2 R + p_3 S \tag{3}$$

Here, L (the total loss) combines three components: quantile-based loss (*Q*), rainy-day loss (*R*), and spatial correlation loss (*S*). These components were weighted with $p_1 = 0.99$, $p_2 = 0.01$ and $p_3 = 1$ to constrain the values within similar ranges. This formulation was designed to ensure overall distributional similarity while also paying attention to spatial structures and rainy-day occurrence during model training. A description of individual loss components follows:

$$Q = \frac{1}{K} \sum_{k=1}^{K} g(q_k) \cdot |Q_X(q_k) - Q_Y(q_k)| \tag{4}$$

Here $Q_X(q_k)$ is the value of the empirical quantile function (inverse CDF) of data X at quantile level $q_k$; $Q_Y(q_k)$ represents the same but for reference data Y. *K* is the total number of quantile levels used (1000), and $q_k$ refers to the uniformly-spaced quantile levels. The weights $g(q_k)$ allow emphasis on specific quantiles; we used the kernel function

$$g(q_k) = \exp(-|q_k - q^*|) \tag{5}$$

where $q^*$ is the quantile we want to emphasize. This is particularly useful for precipitation, where it is important to correct bias in the extremes as well as the bulk of the distribution. If no emphasis is specified, then $g(qk) = 1$ (uniform). We treated $q^*$ as a hyperparameter in our training, choosing a value of either 0.5 (bulk of the precipitation distribution) or 0.9 (tail of the distribution) for each GCM based on performance (other values were tested).

The second loss component is the rainy-day loss function, which is a soft constraint formulated to preserve the frequency of wet/dry days in the data. As a note, if this frequency shifts over time, the monotone transformation used should force the model to learn this shift based on the loss feedback.

$$R = \frac{1}{N}\sum_{i=1}^{N}\left|\sum_{t} \sigma(X_{it} - \alpha) - \sum_{t} \sigma(Y_{it} - \alpha)\right| \tag{6}$$

Here, $X_{it}$ and $Y_{it}$ are bias-adjusted and reference data, respectively, for site *i* and time *t*.  is the sigmoid function which provides a differentiable approximation of the indicator function for rainy/wet days.  is the wet day threshold (1 mm). *N* is the number of samples or sites.

The third loss component is the spatial correlation loss, which is a constraint added to preserve spatial structures in the bias-corrected simulation.

$$S = \frac{1}{BT}\sum_{b=1}^{B}\sum_{t=1}^{T}\left(1 - correlation\left(X_{b,t}{}' Y_{b,t}\right)\right) \tag{7}$$

$$correlation\left(X_{b,t}{}' Y_{b,t}\right) = \frac{\sum_{p=1}^{P} Y_{b,p,t} X_{b,p,t}}{\sqrt{\sum_{p=1}^{P} Y_{b,p,t}^{2} + \epsilon}\sqrt{\sum_{p=1}^{P} X_{b,p,t}^{2} + \varepsilon}} \tag{8}$$

Here, $X_{b,t}$ and $Y_{b,t}$ are bias-adjusted and reference data, respectively, for batch *b* and time *t*. *B* is the total number of batches, *T* is the total number of time steps, and *P* is the number of neighbors for each of the grid cells in B, with iterator *p*.  is a small constant (e.g., $10^{-8}$) to prevent division by zero.

Because every GCM has a distinctive, characteristic large-scale bias pattern (Krinner & Flanner, 2018), no single bias adjustment transformation is suitable for all GCMs. We therefore performed multiple training runs for each GCM, varying hyperparameters like $q*$ and the number of training epochs. For every candidate configuration, we first checked whether the quantile loss was monotonically decreasing over the epochs, and then calculated the ETCCDI indices (described in section 2.4) for the validation period. Next, we computed a composite score defined as the mean percentage bias across all indices and selected the hyperparameter set with the lowest composite score for that GCM.

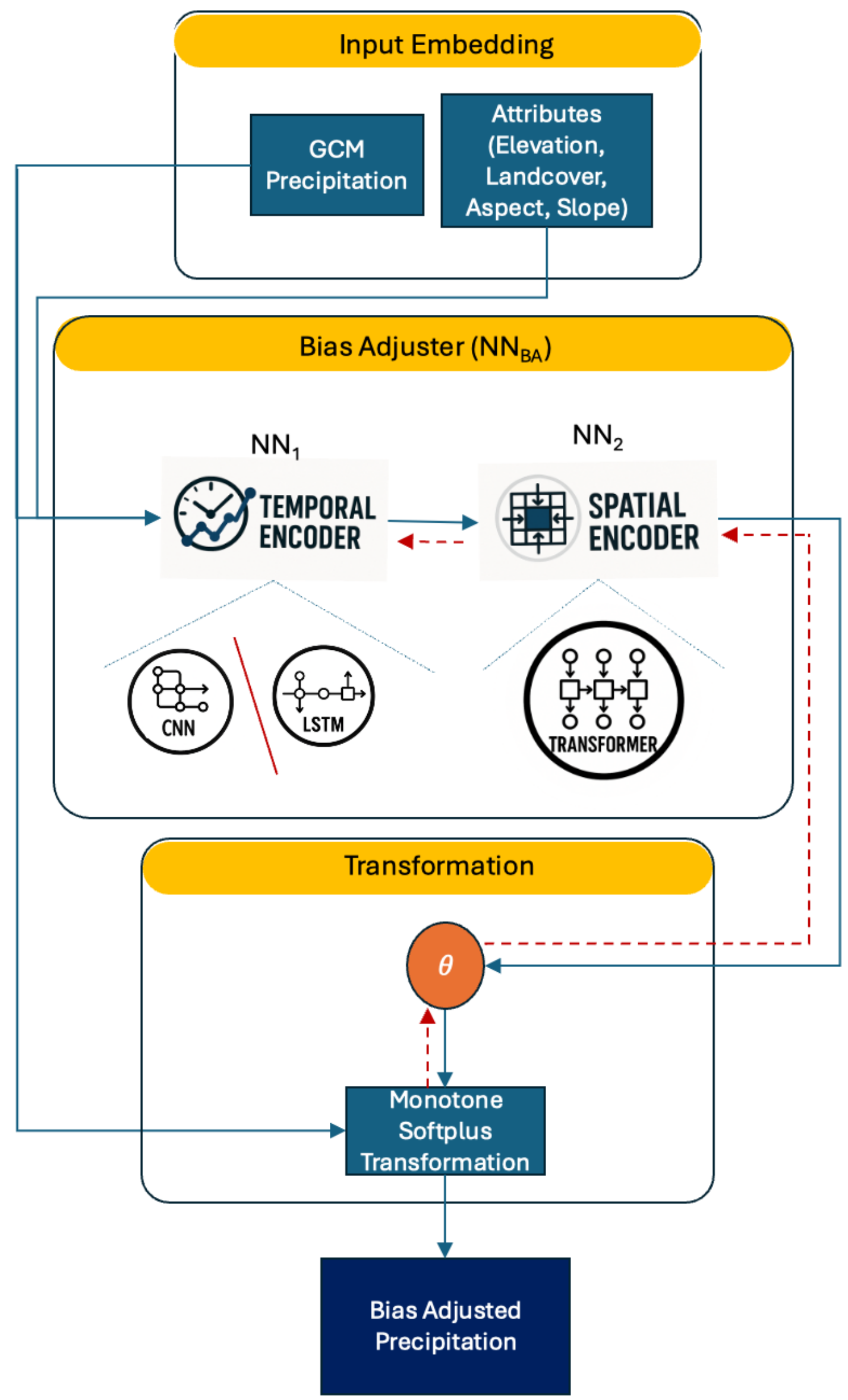

**Figure 1: Schematic of the δCLIMBA framework for differentiable precipitation bias adjustment. Raw General Circulation Model (GCM) precipitation and static spatial attributes (elevation, landcover, aspect, slope) are embedded and passed to a bias-adjustment network ($NN_{BA}$). The network consists of temporal and spatial encoders which jointly estimate parameters θ governing a monotonic transformation (f(x)). The transformation module applies a learnable, differentiable monotone mapping to produce bias-adjusted precipitation based on the feedback from the composite quantile loss function.**

### 2.2 Datasets

We employed daily precipitation data from Coupled Model Intercomparison Project Phase 6 (CMIP6) (Eyring et al., 2016b) model outputs from historical experiments and future scenarios with high emissions and stronger reliance on fossil fuels i.e., Shared Socioeconomic Pathways (SSP5-8.5). For the current scope, the following six CMIP6 models were selected: ACCESS-CM2, GFDL-ESM4, IPSL-CM6A-LR, MIROC6, MPI-ESM1-2-LR, and MRI-ESM2-0 for bias correction of their precipitation. Each of these models was used at their native resolution (1° to 2.5°; see Table 1) within the conterminous U.S. (CONUS) region at the daily timescale for historical simulation.

We considered using gridded Livneh (Livneh et al., 2013) data as a reference for comprehensive evaluation of our model over the entire CONUS. However, we used a revised version called unsplit-Livneh (Pierce et al., 2021) for training our model since it eliminates the problematic time adjustment applied in the original dataset and was also used for calibration of LOCA2. To enable comparison at GCM scale, the unsplit Livneh dataset was aggregated (spatially) from the original resolution of 1/16° to the coarser resolution of the respective CMIP6 members via nearest-neighbor interpolation.

The Localized Constructed Analogs (LOCA2) dataset (Pierce et al., 2014, 2021) is considered as a very important resource for downscaled products of CMIP6 (Vano et al., 2020). It presents a downscaled product with substantially improved resolution of 1/16° grid as it was calibrated using unsplit-Livneh. LOCA2 leverages spatial analog approaches to match patterns from regional- to local-scale features. Due to its ubiquity in our target CONUS region and thorough vetting (Ullrich, 2023), we decided to use it as a benchmark product against our differentiable approach along with other bias correction methods. To use this, we upscaled LOCA2 to GCM's spatial resolution to evaluate purely for systematic biases.

Furthermore, input features like elevation, aspect, slope, and landcover were used to assist precipitation bias adjustment. Elevation, slope, and aspect were extracted from Digital Elevation Models at 0.01° resolution from NASA Shuttle Radar Topography Mission Global 1 arc second V003 dataset (SRTMGL1) (NASA JPL, 2013). Landcover data was produced from Landsat at 30m US coverage and downloaded from CEC's North American Land Change Monitoring System (NALCMS) (Pasos, 2020). All these data features were regridded to respective GCMs from CMIP6 to match the domain.

**Table 1: Description of General Circulation Models (GCMs)**

| **Short Name** | **Full Name** | **Institution/consortium** | **Native Resolution** |
|---|---|---|---|
| ACCESS-CM2 | Australian Community Climate and Earth System Model version 2 | CSIRO / Bureau of Meteorology / Australian universities (ACCESS community) (Bi et al., 2020) | ~1.25$^{\circ}$x ~1.87$^{\circ}$ |
| GFDL-ESM4 | Geophysical Fluid Dynamics Laboratory Earth System Model 4 | NOAA Geophysical Fluid Dynamics Laboratory (NOAA-GFDL) (Krasting et al., 2018) | ~1.00$^{\circ}$x ~1.25$^{\circ}$ |
| IPSL-CM6A-LR | Institut Pierre-Simon Laplace Climate Model version 6A, low resolution | Institut Pierre-Simon Laplace (IPSL), France (Boucher et al., 2018) | ~1.26$^{\circ}$x ~2.50$^{\circ}$ |
| MIROC6 | Model for Interdisciplinary Research on Climate version 6 | MIROC consortium: JAMSTEC, AORI (Univ. Tokyo), NIES, R-CCS Kobe, others (Mochizuki et al., 2019) | ~1.40$^{\circ}$x ~1.40$^{\circ}$ |
| MPI-ESM1-2-LR | Max Planck Institute Earth System Model version 1.2, low resolution | Max Planck Institute for Meteorology (MPI-M), Germany (Gutjahr et al., 2019) | ~1.87$^{\circ}$x ~1.87$^{\circ}$ |
| MRI-ESM2-0 | Meteorological Research Institute Earth System Model version 2.0 | Meteorological Research Institute (MRI), Japan (Kawai et al., 2019) | ~1.12$^{\circ}$x ~1.12$^{\circ}$ |

### 2.3 Experiment Design

We trained the model on the period 1979–2000 and validated it on the period 1965–1978. We then tested bias-corrected outputs of the GCMs over the period 2001–2014. For the evaluation of future scenarios, we chose Shared Economic Pathways (SSP5-8.5) (Intergovernmental Panel On Climate Change (Ipcc), 2023) for the period 2015–2099. We interpolated the reference data (unsplit-Livneh) using nearest-neighbor interpolation to the resolution of each GCM before using it for modeling. We evaluated the model by comparing it to a set of five bias-correction methodologies (Table 2).

We performed our experiments on the Perlmutter system at the National Energy Research Scientific Computing Center (NERSC). Training our model on six GCM members across the CONUS for 100 epochs on different hyperparameter combinations distributed over 4 A100 GPUs in parallel took 7 hours. After training, it only took 2 minutes to produce a bias-corrected simulation across the whole CONUS for the testing and future periods.

**Table 2: Summary of the bias adjustment methods used for evaluation. All methods except LOCA2 (reference comparison) were implemented using the 'ibicus' framework proposed in Spuler et al. (2024).**

| **Method** | **Description** |
|---|---|
| Quantile Mapping (QM) | Statistical bias correction that adjusts the modeled distribution so its quantiles match those of observations (Cannon et al., 2015; Maraun, 2016). |
| ISIMIP (ISIMIP3BASD) | Trend-preserving bias adjustment and statistical downscaling framework developed for the Inter-Sectoral Impact Model Intercomparison Project; uses parametric/nonparametric quantile mapping tailored to each variable, with explicit preservation of trends across quantiles and options for multivariate/space–time structure (Lange, 2019). |
| ECDFM | Nonparametric quantile mapping where empirical CDFs are used and the correction is applied as an additive (or multiplicative) "distance" between modeled and observed quantiles (H. Li et al., 2010; L. Wang & Chen, 2014). |
| Quantile Delta Mapping (QDM) | Bias-correction method that corrects biases at each quantile while explicitly preserving the modeled climate-change signal at that quantile (Cannon et al., 2015; L. Wang & Chen, 2014). |
| LOCA2 (upscaled) | Updated hybrid statistical downscaling method that extends LOCA: uses an analog-based constructed-analog framework with observational training data, two-step coarse-to-fine downscaling, and explicit seasonal, intensity-dependent bias correction (Pierce et al., 2021). |

### 2.4 Evaluation Design and Metrics

To evaluate our model, we focused on investigating the characteristics of bias-corrected simulations through Expert Team on Climate Change Detection and Indices (ETCCDI) for marginal biases, fractal dimension for spatial patterns/clustering, and trend bias percentage.

To evaluate the temporal structure and marginal biases of the corrected data, we calculated the ETCCDI indices listed in Table 3. To place the different GCMs and bias adjustment methods on an equal footing during the evaluation, we first interpolated all of the bias-adjusted simulations to the resolution of the coarsest GCM (IPSL-CM6A-LR, $1.25^{o}$x$2.5^{o}$).

**Table 3: Summary of ETCCDI Indices for precipitation evaluation**

| **ETCCDI Indices** | **Definition** |
|---|---|
| R10mm | Annual count of days when precipitation ≥ 10mm |
| R20mm | Annual count of days when precipitation ≥ 20mm |
| Rx1day | Monthly maximum 1-day precipitation amount |
| Rx5day | Monthly maximum 5-day precipitation amount |
| Simple Daily Intensity Index (SDII) | Monthly average precipitation amount on wet days (precipitation ≥ 1mm) |
| Consecutive Dry Days (CDD) | Annual maximum number of consecutive days with precipitation < 1mm |
| Consecutive Wet Days (CWD) | Annual maximum number of consecutive days with precipitation ≥ 1mm |
| R95pTOT | Annual total precipitation from days exceeding the 95th-percentile of daily precipitation during a reference period |
| R99pTOT | Annual total precipitation from days exceeding the 99th-percentile of daily precipitation during a reference period |

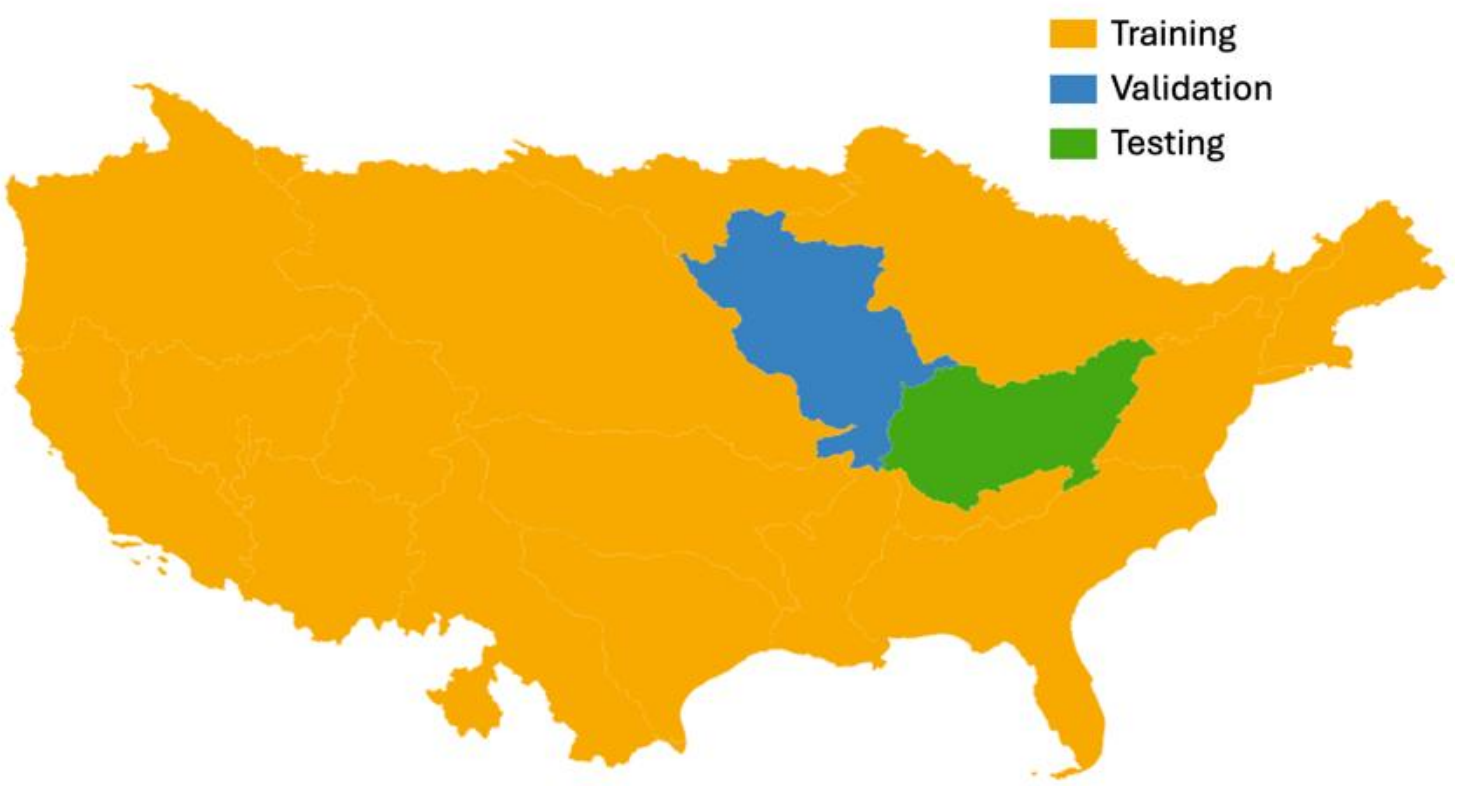

**Figure 2: Spatial configuration of training, validation, and testing domains used to evaluate δCLIMBA in data-scarce settings. Validation and testing regions were entirely withheld from model training.**

To evaluate spatial structure, we used the scale-dependent fractal dimension (FD) computed via the box-counting method (Husain et al., 2021; Lovejoy et al., 1987; Meisel et al., 1992) applied to thresholded binary precipitation fields at each quantile level. This method has been used to study how precipitation patterns change across scales (Hess et al., 2023; Lovejoy et al., 1987). For a given quantile threshold $h,$ the precipitation field $J$ is converted to a binary mask $M_h$:

$$M_h = \{1 \text{ if } \geq \text{ J else } 0\} \tag{9}$$

To measure how this thresholded pattern occupies space across scales, the mask is then partitioned using square boxes of side length ℓ. For each ℓ, we count the number of partially filled boxes, $\mathcal{N}(\ell)$, i.e., boxes that are neither empty nor completely full. If fractal scaling is assumed to be

$$\mathcal{N}_\ell \propto \ell^{-FD} \tag{10}$$

then

$$FD = \frac{\log(\mathcal{N}_\ell)}{\log(1/l)} \tag{11}$$

For evaluating changes in climate signal trend, we used the framework from the 'ibicus' package (Spuler et al., 2024). For a particular GCM and bias-correction method, we first computed the raw trend ($\mathcal{T}_{\text{raw}}$) for a statistic $\mathcal{S}$ (e.g., mean or 95th quantile) between the future (2015–2099) and historical (1980–2014) periods:

$$\mathcal{T}_{\text{raw}} = \mathcal{S}_{raw,future} - \mathcal{S}_{raw,\text{historical}} \tag{12}$$

Then, we calculated the trend after carrying out bias adjustment:

$$\mathcal{T}_{debiased} = \mathcal{S}_{debiased,future} - \mathcal{S}_{\text{debiased,historical}} \tag{13}$$

Finally, the trend bias ($TB$) was calculated as the percentage change between $\mathcal{T}_{\text{raw}}$ and $\mathcal{T}_{debiased}$:

$$\text{TB} = 100\ \frac{\mathcal{T}_{debiased} - \mathcal{T}_{\text{raw}}}{\mathcal{T}_{\text{raw}}} \tag{14}$$

One of the advantages of this differentiable framework is its ability to generalize not only in time but in space, provided we ingest data that helps it to learn spatial context (e.g., elevation, slope, and landcover). Our final evaluation was a spatial test where we trained, validated, and tested the model on different regions (Fig. 2). We trained and evaluated the model on the period 1990–2014, holding out the Upper Mississippi region (blue) for validation to select hyperparameters, and used this learned configuration to test in the adjacent Ohio region (green).

## 3. Results

δCLIMBA was evaluated using a series of experiments to test its generalizability across different GCMs at different locations, including its ability to adjust biases through precipitation indices, conserve spatial structures, and capture long-term trends.

### 3.1 Quantile Comparison

Quantile comparisons across five climatically diverse U.S. cities demonstrate that δCLIMBA systematically aligns the daily precipitation distributions of all six CMIP6 models with the Livneh reference across the full distributional range, including the upper tail, without inducing the nonphysical amplification of extremes that is characteristic of several conventional adjustment methods (Fig. 3). The five cities, Phoenix, Yosemite, Seattle, Orlando, and Philadelphia respectively span arid, mountainous, maritime, humid subtropical, and temperate climatic regimes.

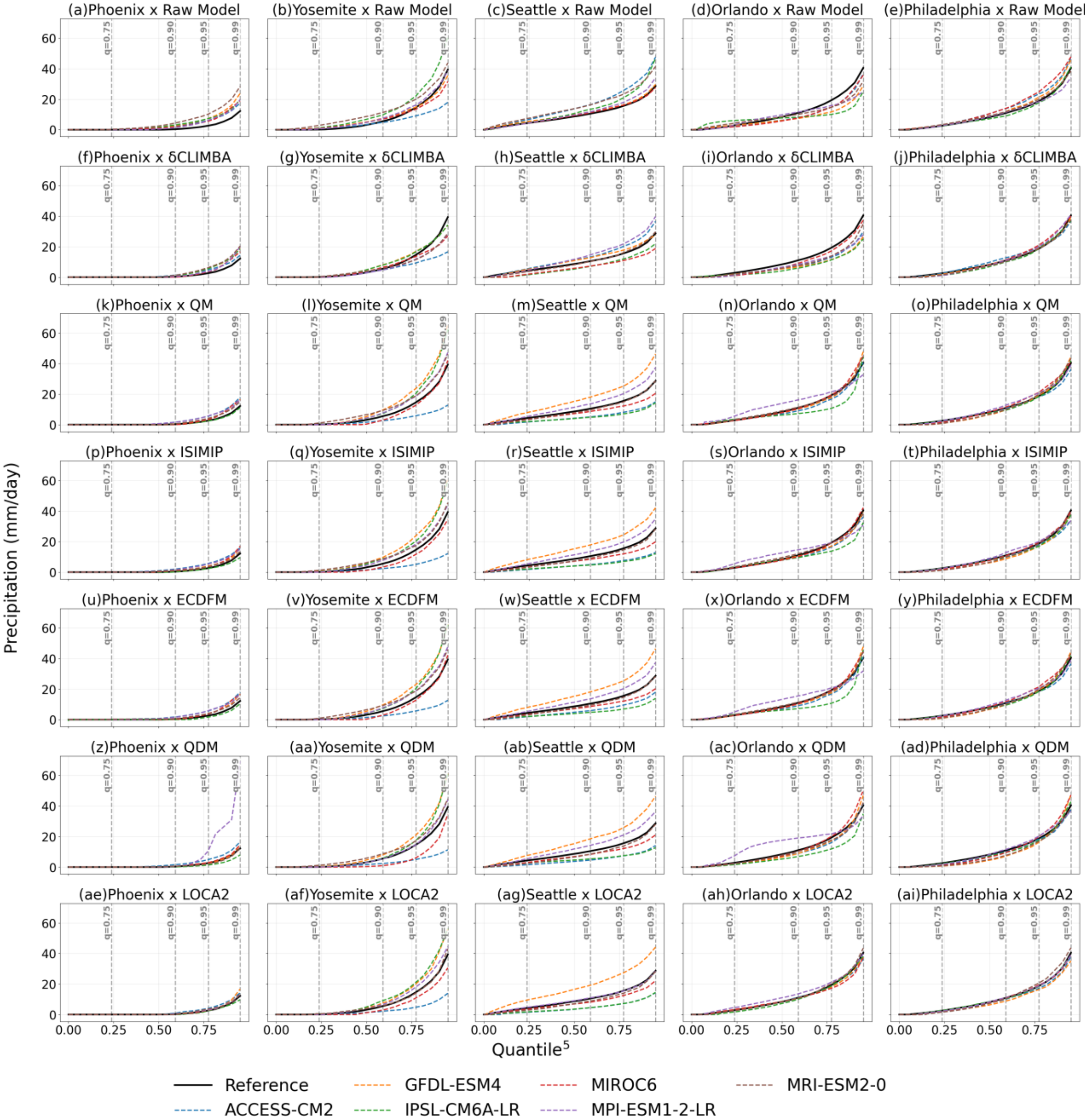

**Figure 3: Quantile comparison of daily precipitation across selected climatically-diverse cities and bias adjustment methods during the testing period (2001–2014). For each city–method combination, empirical quantile curves of the six CMIP6 General Circulation Models are shown alongside the Livneh reference distribution. The x-axis displays quantiles to the 5$^{th}$ power to expand the upper tail and emphasize extreme precipitation behavior (the original quantiles are marked using vertical dashed lines across each plot). The y-axis represents precipitation (mm day$^{-1}$). Columns correspond with different cities and rows with bias adjustment methods**.

While the raw models exhibited large spread and biases compared to observations, δCLIMBA consistently aligned model quantiles with the reference across the full distribution while preserving relative inter-model differences and avoiding artificial tail inflation (Fig. 3). The raw models' large inter-model spread and systematic biases were strongly regime-dependent: most models generally matched observations in Philadelphia, but displayed dry-region overestimation in Phoenix, exaggerated tails in mountainous Yosemite, and inconsistent heavy-rain behavior in humid Seattle and Orlando. Conventional bias-correction methods reduce mean bias but frequently distort tail behavior: Quantile Mapping, ISIMIP, and ECDFM often over-amplified upper quantiles, while Quantile Delta Mapping produced unstable and occasionally completely unrealistic tails (most evident in Phoenix and Orlando). LOCA2 improved inter-model consistency but still showed city-dependent residual biases in extremes. δCLIMBA performed well across multiple regimes and GCMs, indicating that the differentiable, spatially-conditioned formulation corrected distributional bias without inducing non-physical extreme amplification, a failure mode clearly visible in several traditional methods.

### 3.2 Precipitation Indices

Across nearly all indices, including intensity (Rx1day, Rx5day), threshold exceedance (R10mm, R20mm), persistence (CDD, CWD), and extreme totals (R95pTOT, R99pTOT), δCLIMBA exhibited median bias closer to zero compared to traditional statistical methods (Fig. 4). In contrast, raw model output and several conventional corrections showed large positive biases for extreme and heavy-precipitation indices, as well as substantial dispersion, highlighting their limited ability to jointly control mean bias and spatial consistency.

While Fig. 4 summarizes the distribution of grid-wise biases, it does not reveal whether residual errors exhibit coherent spatial structure. Figure 5 therefore examines the spatial distribution of mean percentage bias across CONUS for selected ETCCDI indices. This analysis allows assessment of whether δCLIMBA merely reduces aggregate bias or also mitigates geographically systematic error patterns.

Across all metrics except CDD, the raw GCM output exhibited widespread positive bias over the CONUS (Fig. 5). While conventional bias adjustment methods such as Quantile Mapping (QM) effectively reduced this bias, their performance varied substantially across regions. For instance, QM retained a positive residual bias in the southeastern United States relative to both LOCA2 and δCLIMBA for multiple extreme metrics including R99pTOT, SDII, Rx5day, and CWD.

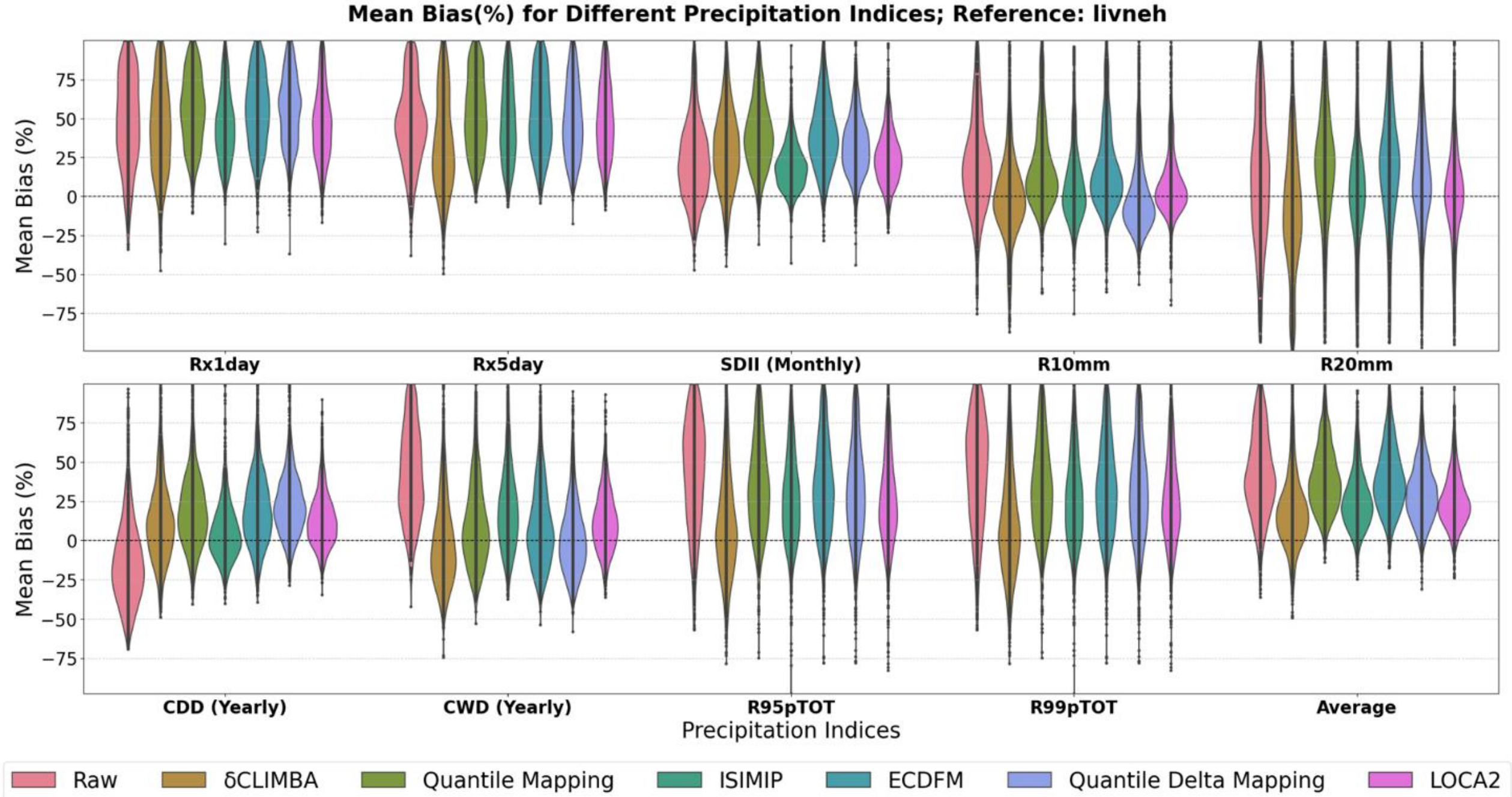

**Figure 4: Spatial distribution of mean percentage bias for ETCCDI precipitation indices over the conterminous U.S. (CONUS) during the test period (2001–2014), evaluated against the Livneh reference dataset. Results are shown for the ensemble of six CMIP6 General Circulation Models. Each violin represents the distribution of grid-point mean percentage bias across the CONUS for a given index and method. The "Average" column reports the arithmetic mean of the mean percentage bias across all ETCCDI indices for each method. Narrower distributions and medians (widest point in violin) closer to zero indicate improved bias stability and reduced spatial heterogeneity.**

For extreme precipitation totals (R99pTOT), QM exhibited a pronounced positive bias across the entire region. In contrast, δCLIMBA mitigated these extremes without introducing systematic negative compensation elsewhere, indicating that its learned monotonic transformation adapted spatially. This behavior aligns with the spatial conditioning embedded in the neural architecture which enables the model to gain spatial context while attenuating bias. In addition to that, the weighted quantile loss function (assigning a higher penalty to higher quantiles) gives direct feedback to the model to de-bias precipitation at the higher end of the distribution. While LOCA2 achieved a comparable performance to δCLIMBA, it retained a moderately higher residual bias across the CONUS.

Intensity and frequency-based indices (SDII and R20) were challenging for all approaches. For SDII, bias adjustment techniques counterintuitively increase bias relative to raw CMIP6 output rather than reducing it (Fig. 5). This degradation was most pronounced in the central United States, where QM introduced the largest bias amplification, while LOCA2 and δCLIMBA exhibited comparable residual bias. The threshold-based index R20 presented a distinct challenge: unlike other metrics, R20 exhibited systematic underestimation, particularly across the western mountainous region. Nevertheless, δCLIMBA and LOCA2 successfully reduced positive bias of R20 at many locations, suggesting that the monotonic transformation constraint preferentially suppresses high-frequency precipitation events.

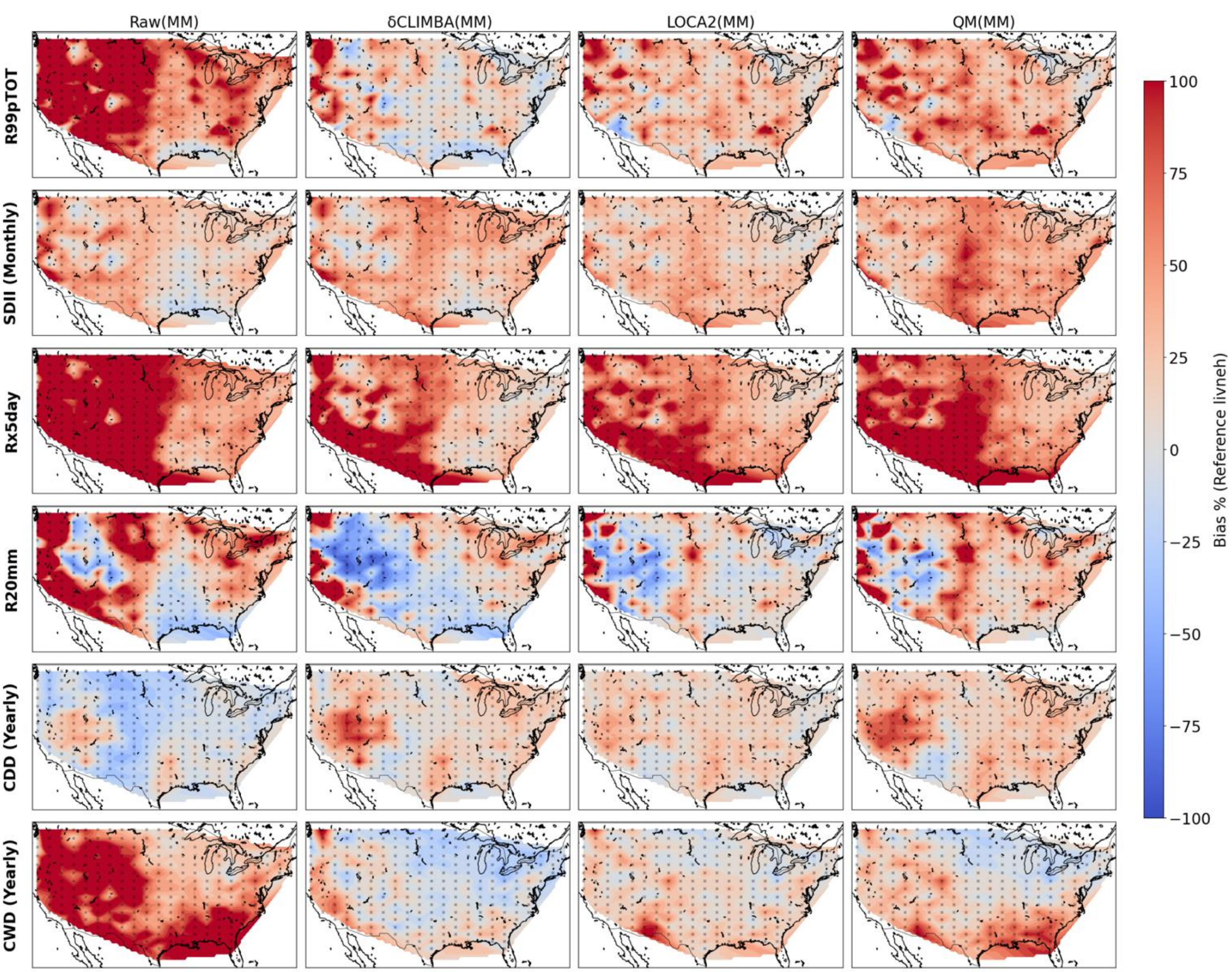

**Figure 5: Spatial distribution of mean percentage bias (%) for selected ETCCDI precipitation indices over the conterminous U.S. during the test period (2001–2014), evaluated against the Livneh reference dataset. Results are shown for the ensemble mean (model mean, MM) of six CMIP6 General Circulation Models. Rows correspond to precipitation indices, and columns correspond to bias-adjustment methods. Positive values indicate overestimation relative to Livneh, and negative values indicate underestimation. This figure highlights the spatial structure of residual bias, complementing the distributional summaries shown in Fig. 3.**

All approaches achieved substantial bias reduction for both temporal sequencing metrics (CDD and CWD) across the CONUS. For CWD, positive bias was effectively attenuated throughout most of the domain, with persistent residual bias limited to a few West Coast locations. Both δCLIMBA and LOCA2 outperformed QM in the high-rainfall southeastern region. For CDD, negative bias was similarly reduced across most locations, though a cluster of grid points in the western interior retained positive residual bias across all methods.

Collectively, these spatial diagnostics revealed heterogeneous performance of δCLIMBA across different precipitation characteristics. The method achieved consistent improvements in extreme totals (R99pTOT) and temporal sequencing metrics (CDD, CWD), while matching or exceeding LOCA2's performance and substantially outperforming conventional QM. Since GCMs usually tend to carry systematic wet (positive) biases (Srivastava et al., 2020), the transformation in δCLIMBA seems particularly amenable to mitigating positive bias; especially in regions like the southeastern US where precipitation distribution is highly skewed and frequency and intensity is driven by mesoscale convective systems (Rahman et al., 2023). However, intensity-based indices (SDII, R20) remain challenging for all bias correction approaches, highlighting the need for methods that can explicitly account for changes in precipitation event structure.

### 3.3 Spatial Patterns

δCLIMBA substantially preserved the multiscale spatial structure of precipitation fields, achieving a mean absolute error (MAE) of 0.022 relative to the Livneh reference fractal dimension curve, a marked improvement over raw CMIP6 output (MAE=0.030) and approaching the performance of LOCA2 (MAE=0.017) as shown in Fig. 6. This result is notable because conventional univariate bias adjustment methods, operating independently at each grid point, lack any mechanism for learning spatial relationships and therefore tend to attenuate marginal bias at the cost of degrading spatial organization (Allard et al., 2025; Maraun et al., 2017). The fact that δCLIMBA achieves competitive spatial fidelity without an explicit analog procedure suggests that geodesic-aware spatial attention in the spatial encoder is capturing meaningful inter-node structure during training.

While LOCA2 achieved the lowest overall MAE (0.017), δCLIMBA competitively tracked the reference fractal-dimension curve across nearly the entire quantile range, with substantially reduced error relative to the raw model (MAE = 0.022 vs. 0.030). It exhibited good performance across the distribution, except for a deviation in the upper tail that is likely due to numerical instability, indicating that the differentiable framework corrects biases while retaining a physically realistic multiscale spatial structure. LOCA2’s superior MAE is consistent with its design: the spatial analog method explicitly maps regional precipitation patterns to local scales and thus inherently reproduces aspects of spatial intermittency as a direct consequence of its construction, rather than as an emergent property of learned representations.

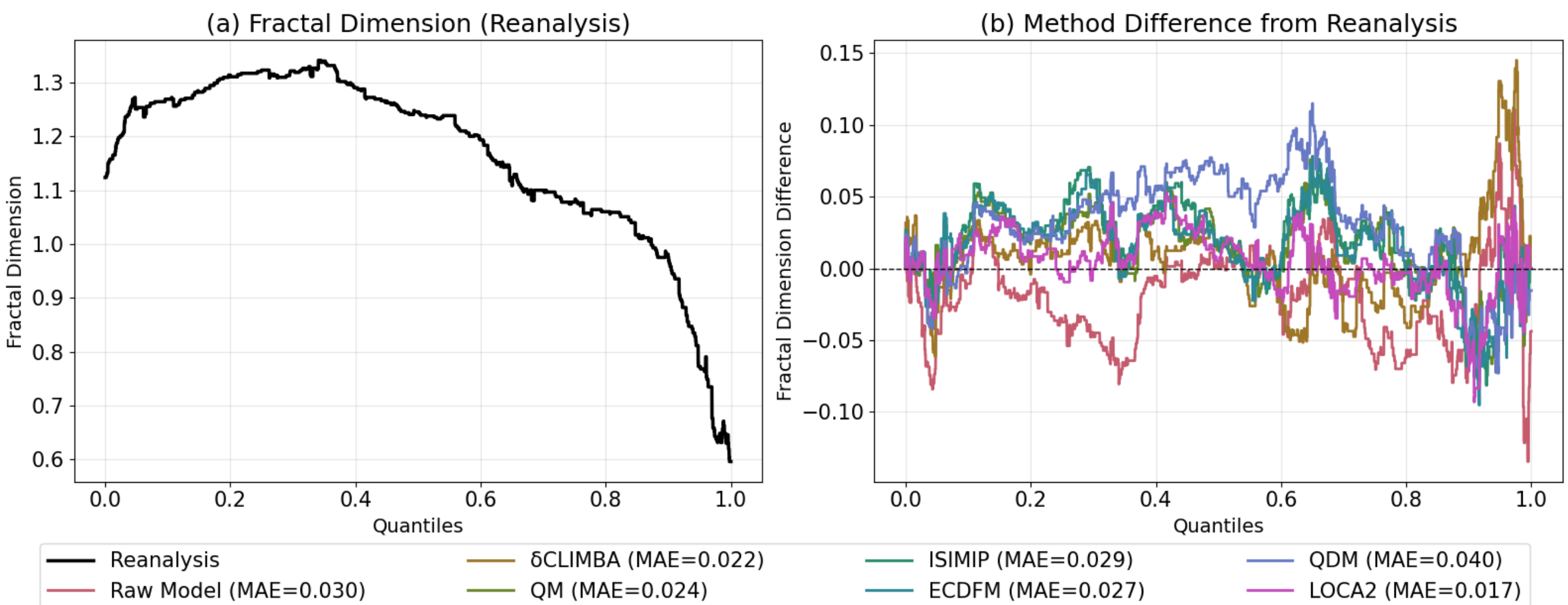

**Figure 6: Fractal dimension of precipitation fields as a function of quantile over the conterminous U.S. during the test period (2001–2014). The black curve represents the Livneh reference dataset, while colored curves correspond to ensemble model mean (MM) fields after bias adjustment (δCLIMBA, LOCA2, Quantile Delta Mapping (QDM)) and raw General Circulation Model output. Mean absolute error (MAE) values reported in the legend quantify deviation from the reference fractal dimension curve across quantiles.**

### 3.4 Partial Trend Preservation

δCLIMBA partially preserved the trend in projected precipitation for GFDL-ESM4 under SSP5-8.5 without any explicit trend-preservation mechanism embedded in the framework, outperforming QDM and LOCA2 across all evaluated metrics while trailing ECDFM and ISIMIP for mean precipitation and the 95th percentile of daily precipitation. However, this behavior did not generalize consistently across GCMs, indicating that trend preservation is an emergent rather than a structural property of the current framework and demands more future investigation.

Figure 7 shows the bias in projected precipitation trends between the historical (1970-2014) and future periods (2015-2099) under SSP5-8.5 for mean precipitation, 95th percentile of daily precipitation, wet days (> 1 mm day$^{-1}$), and very wet days (> 10 mm day$^{-1}$) across multiple bias-adjustment methods. Quantile Delta Mapping (QDM) and LOCA2 (resolution upscaled) exhibited pronounced negative trend biases across all metrics, indicating systematic attenuation of trends after bias correction. Quantile Mapping performed better than its counterparts in preserving the trend of mean and 95th percentile daily precipitation, lagged behind for wet days, and performed comparably for trend in very wet days (>10mm).

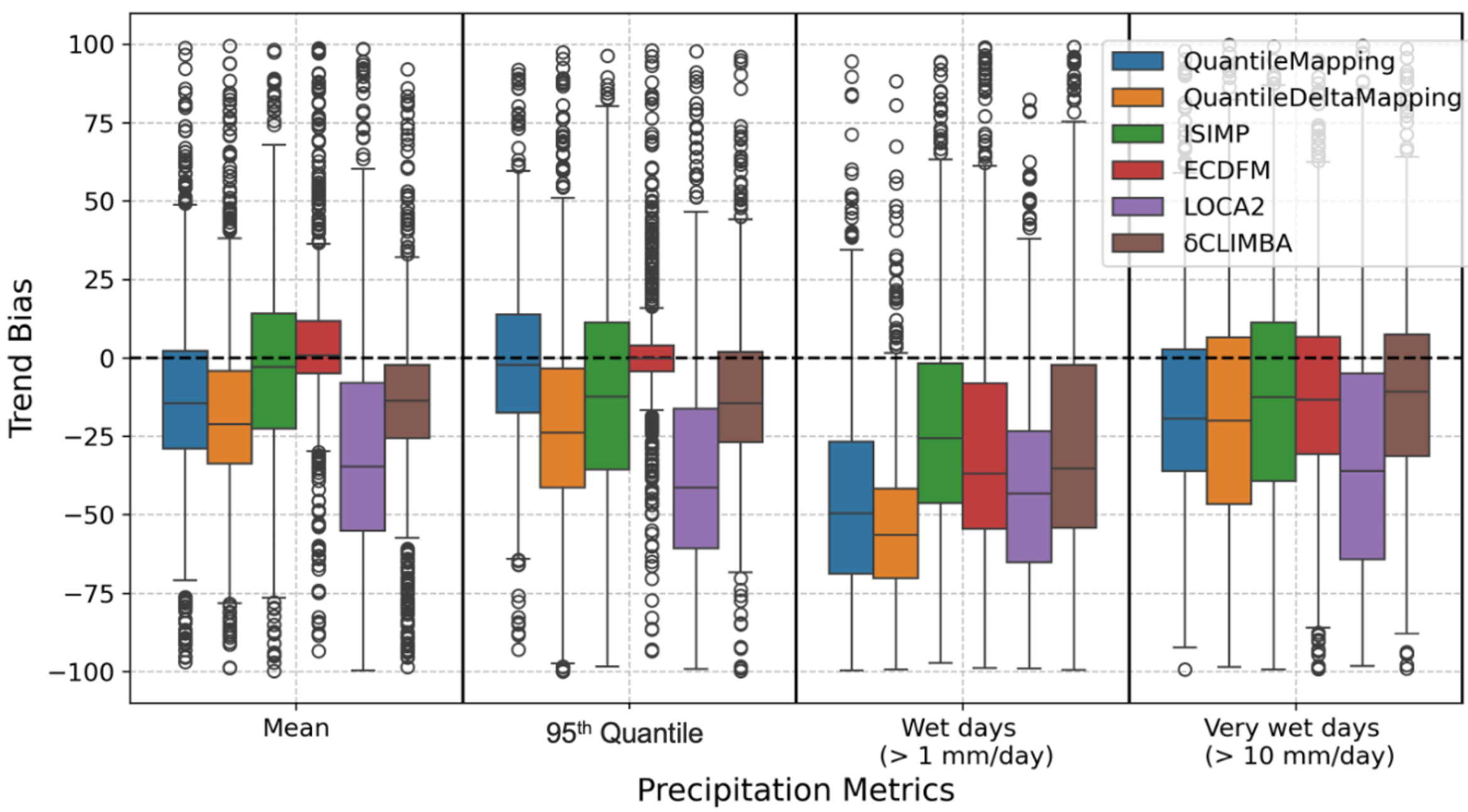

**Figure 7: Percentage bias in projected climate trends of GFDL-ESM4 between future and historical periods under SSP5-8.5, evaluated across precipitation metrics over CONUS. Values closer to zero indicate improved preservation of trends following bias adjustment. The central line on each box indicates the median, the top and bottom of each box respectively indicate the 25th and 75th percentiles, and the top and bottom whiskers of each box respectively indicate the 2.5th and 97.5th percentiles. The open dots indicate outliers.**

ECDFM, which is well recognized for its trend preservation (H. Li et al., 2010), did a particularly good job with trend bias in the mean and 95th percentile of daily precipitation, although its performance was more typical when evaluated for trends in number of wet days and very wet days. ISIMIP, another bias-correction method designed for trend-preservation, displayed similar performance to ECDFM, though with a higher inter-quartile range, implying a larger number of locations with distorted trends.

δCLIMBA has no built-in trend-preservation mechanisms, but shows acceptable performance, nonetheless. For mean and 95th percentile of daily precipitation, it performed similarly to the two quantile-mapping methods, lagging behind ISIMIP and ECDFM but doing better than LOCA2, while for wet and very wet days, its performance was competitive with ISIMIP and ECDFM. It exhibited similar performance on some other GCMs, like ACCESS-CM2, but not all, indicating that the differentiable framework was unable to generalize the trend-preservation behavior across GCMs. Future δCLIMBA versions could investigate loss terms focusing on long-term trends.

### 3.5 Bias Correction in Data Scarce Regions

δCLIMBA demonstrated meaningful spatial generalization to an entirely withheld region, reducing positive bias in extreme precipitation indices including R99pTOT and Rx5day without any local calibration except for validation (Fig. 8). However, indices sensitive to daily wet-day intensity (SDII, R20mm) exhibited residual overestimation, indicating a limit on the spatial transferability of the current framework.

Improvements were most pronounced for intensity-based and percentile-based indices, including Rx1day, Rx5day, and high-percentile precipitation (R95pTOT and R99pTOT) totals, indicating that the learned monotonic transformation effectively corrected systematic distributional distortions in unseen regions. Some indices, such as SDII, CDD and CWD, retained wider bias distributions, suggesting that duration-based metrics remain more challenging to correct under spatial holdout. However, the longer interquartile range in the violin plots for all indices shows that many of the grid cells in the test region tend to end up with biases even larger than with the raw simulation, such as SDII, R10, and CDD.

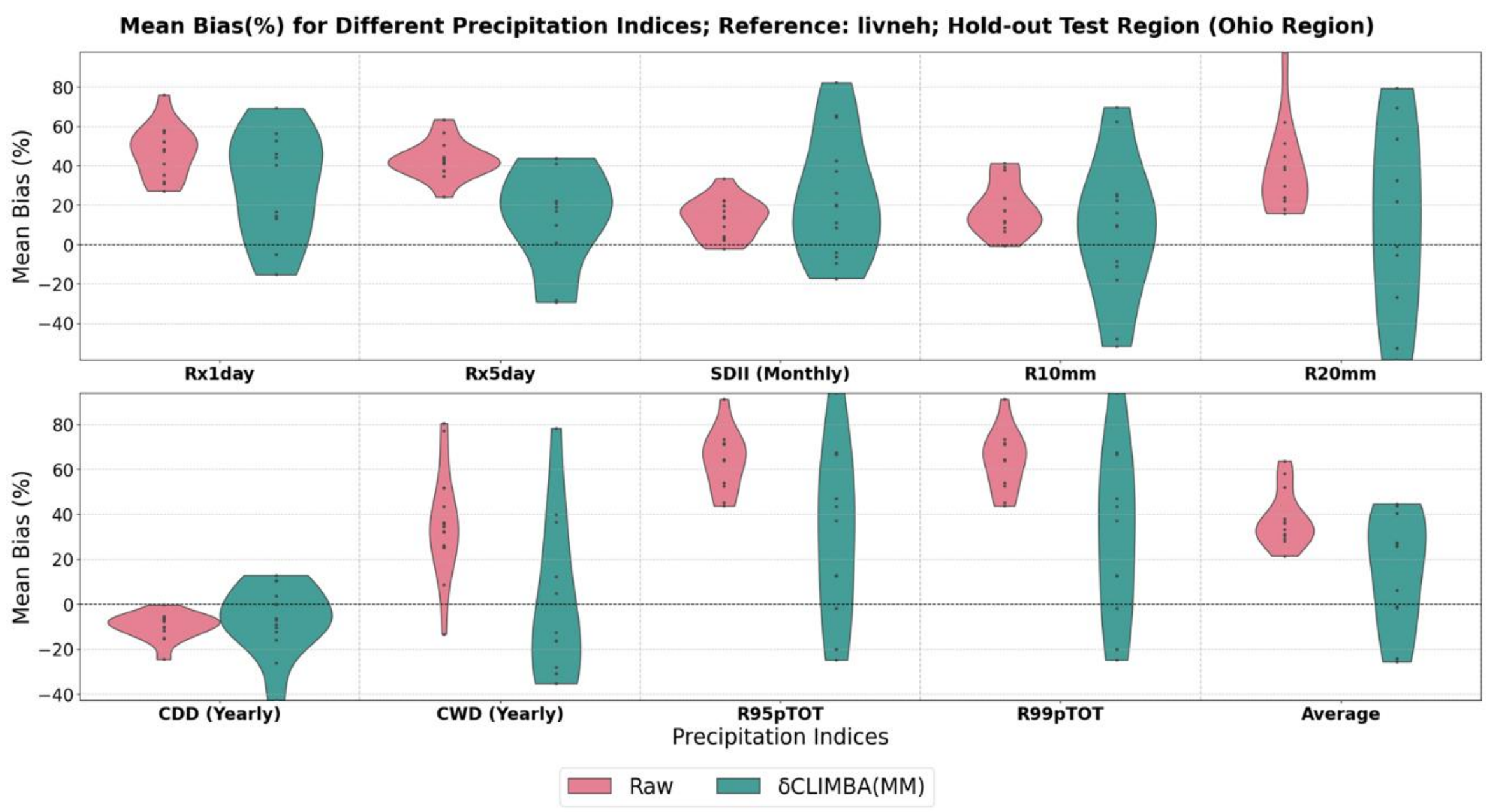

**Figure 8: Spatial Distribution of mean percentage bias for ETCCDI precipitation indices over the testing region (Ohio region), evaluated against the Livneh reference dataset. Results are shown for the ensemble mean (model mean, MM) of six CMIP6 General Circulation Models. Each violin represents the distribution of grid-point mean percentage bias across the testing region for a given index and method. The "Average" column reports the arithmetic mean of mean percentage bias across all ETCCDI indices. Narrower distributions and medians closer to zero indicate improved bias stability and reduced spatial heterogeneity.**

This experiment constitutes a particularly stringent test of the framework, as conventional bias-correction methods, whether distribution-based, moment-scaling, or spatial analog are explicitly calibrated against local observations and lack a principled mechanism for spatial generalization; their performance inherently degrades outside the calibration domain, making a like-for-like spatial comparison uninformative. In addition, LOCA2 is a bespoke product (Pierce et al., 2014, 2021), so there is no way to test its ability to generalize spatially.

Overall, these results demonstrate that there is promise in using δCLIMBA to generalize spatially. Since the grid cells are moving towards zero bias, the differentiable framework has the potential to maintain stable bias-adjustment behavior when transferred to an unseen but adjacent region, reducing distributional biases without access to local calibration data. Yet, more effort is needed to better regularize the system.

## 4. Conclusions

Bias correction or bias adjustment is a domain that cannot be addressed or optimized by a single metric. Every GCM output carries different kinds of biases, and therefore an exhaustive evaluation of not only marginal biases but their temporal, spatial, and scale-dependent inherent characteristics is important. With its composite quantile loss alone, δCLIMBA excels in adjusting the distributions of GCMs across diverse cities, surpassing all traditional methods while being on par with LOCA2. This similar or even better performance is also depicted while attenuating marginal biases of extreme precipitation indices like R95pTOT and R99pTOT. However, biases in frequency and intensity related indices remain challenging, indicating the monotonic transformation may not be able to efficiently mitigate bias at higher quantiles even though annual precipitation is matched. This could mean that δCLIMBA is also inflating the tail of wet-day intensity distribution relative to the reference dataset, i.e., mapping too much probability mass into higher quantiles. We encourage future work with differentiable models for addressing intensity bias.

Since precipitation is a multiscale phenomenon, δCLIMBA uses a transformer-based spatial encoder to learn spatial structures from neighboring grid cells. Though the current framework is able to only learn short-range dependencies through each grid cell's 16 closest correlated neighbors, δCLIMBA outputs were able to imitate spatial patterns (based on fractal dimension) closest to the reference datasets, second to LOCA2. This framework could be a gateway for more work in adding mechanisms that learn long-range dependencies to output more realistic precipitation simulations.

Even though δCLIMBA performed well on reducing distributional biases and conserving spatial patterns, one of the biggest concerns from every bias correction method is about preservation of trends between historical and future simulations. Unlike ECDFM and ISIMIP, δCLIMBA does not include an explicit trend-preservation mechanism, but still exhibited better behavior than purely quantile-based methods and LOCA2 in the case of GFDL-ESM4. However, this behavior was not repeated with other GCMs, indicating the need for a data-driven mechanism in δCLIMBA to preserve trends. One of the advantages of machine learning based approaches (including differentiable models) is that the trained models are applicable to completely unseen regions and timeframes. Unlike all the traditional methods including LOCA2, δCLIMBA utilizes information from inputs like elevation and landcover to mitigate biases, which allow it to extrapolate those learned patterns in unseen but similar regions. Marginal biases in the testing region such as R99pTOT and Rx5day were demonstrated to be attenuated with their positive bias distribution moving towards zero. However, the larger interquartile range and biases exacerbated in indices like SDII and R20 indicated overestimation of precipitation per day, limiting the model's reliability to gauge intensity and frequency of events.

Overall, the differentiable model is learning the aspects of bias correction on which it is optimized. The bias adjustor paired with a composite quantile loss function makes sure that the bias correction distribution and spatial patterns remain closer to the observation than traditional methods simulations. Though the framework is generalizable, it is not a universal solution designed to attenuate bias in any GCM, but rather provides a framework to train and tune our choice of GCM. Extending this framework using a foundational model approach for one-shot bias correction is one possible way to make it truly generalizable. This differentiable framework could also be leveraged to understand the mitigation of biases directly by analyzing the coefficients generated from neural networks. All in all, the differentiable framework has the potential to isolate and attenuate bias based on the mechanisms implemented inside it. Its flexibility makes it suitable for integration into Earth system post-processing pipelines and impact workflows, as it is a modular, computationally efficient, and extensible bias correction approach that is physically informed, scalable, and compatible with both historical and future applications.

**Financial Support**

This work was supported by the U.S. Department of Energy RGCM award DOE DE-SC0016605 and DE-AC52-07NA27344. This material is based upon work supported by the National Center for Atmospheric Research, which is a major facility sponsored by the National Science Foundation under Cooperative Agreement No. 1852977.

**Code and Data Availability**

The source code for δCLIMBA and data processing can be found in this repo: https://github.com/kasProg/dCLIMBA-release under PSU license. All datasets used are publicly available and referenced accordingly. The CMIP6 data was accessed via Copernicus Climate Data Store under Copernicus license: https://doi.org/10.24381/cds.c866074c. The unsplit-Livneh was downloaded from: https://cirrus.ucsd.edu/~pierce/nonsplit_precip/. The land cover was downloaded from: https://www.cec.org/publications/nalcms/. The elevation was extracted and processed from SRTM: https://www.earthdata.nasa.gov/data/catalog/lpcloud-srtmgl1-003. Slope and aspect were derived from elevation using ArcGIS.

**Author Contribution**

CS conceived the initial study design. KS and CS designed the experiment. KS developed the model code and performed the analysis. KS, SM, PL, KL, and CS analysed the results and helped prepare the manuscript.

**Competing Interests**

CS and KL have financial interests in HydroSapient, Inc., a company which could potentially benefit from the results of this research. These interests have been reviewed by The Pennsylvania State University in accordance with its individual conflict of interest policy, for the purpose of maintaining the objectivity and the integrity of research at The Pennsylvania State University.